\documentclass[letterpaper, 10 pt, conference]{ieeeconf}  

\IEEEoverridecommandlockouts                              

\usepackage{graphics} 
\usepackage{graphicx}
\usepackage{epsfig} 
\usepackage{amsmath} 
\usepackage{amssymb}  
\usepackage{eufrak}
\usepackage{cite}
\usepackage{multirow}
\usepackage{multicol}
\usepackage[table]{xcolor}
\usepackage{caption}
\usepackage{color,soul}

\definecolor{light_red}{rgb}{1.0, 0.44, 0.37}
\definecolor{light_green}{rgb}{0.53, 0.66, 0.42}

\newcommand{\Real}{\mathbb R}

\def\Vec#1{\!\!\hbox{$#1$\kern-0.38em\lower0.85em\hbox{$\vec{}\,$}}\,}%
\newcommand{\bbm}{\begin{bmatrix}}
\newcommand{\ebm}{\end{bmatrix}}
\DeclareMathAlphabet{\mbf}{OT1}{ptm}{b}{n}
\newcommand{\mbs}[1]{{\boldsymbol{#1}}}

\title{\LARGE \bf
DeepMEL: Compiling Visual Multi-Experience Localization \\ into a Deep Neural Network
}

\author{Mona Gridseth and Timothy D. Barfoot\thanks{All authors are with the University of Toronto Institute
for Aerospace Studies (UTIAS), 4925 Dufferin St, Ontario, Canada.  \texttt{mona.gridseth@robotics.utias.utoronto.ca, tim.barfoot@utoronto.ca}}
}

\begin{document}

\maketitle
\thispagestyle{empty}
\pagestyle{empty}

\begin{abstract}
Vision-based path following allows robots to autonomously repeat manually taught paths. Stereo Visual Teach and Repeat (VT\&R) \cite{Furgale2010} accomplishes accurate and robust long-range path following in unstructured outdoor environments across changing lighting, weather, and seasons by relying on colour-constant imaging \cite{Paton2015} and multi-experience localization \cite{Paton2016}. We leverage multi-experience VT\&R together with two datasets of outdoor driving on two separate paths spanning different times of day, weather, and seasons to teach a deep neural network to predict relative pose for visual odometry (VO) and for localization with respect to a path. In this paper we run experiments exclusively on datasets to study how the network generalizes across environmental conditions. Based on the results we believe that our system achieves relative pose estimates sufficiently accurate for in-the-loop path following and that it is able to localize radically different conditions against each other directly (i.e. winter to spring and day to night), a capability that our hand-engineered system does not have.
\end{abstract}


\section{INTRODUCTION}
Vision-based path following algorithms have enabled robots to repeat paths autonomously in unstructured and GPS-denied environments. Furgale et al. \cite{Furgale2010} perform accurate metric and long-range path following with their VT\&R system, which relies on a local relative pose map removing the need for global localization. The authors use sparse SURF features \cite{Bay2006} to match images when performing VO and localization. Paton et al. extend VT\&R to autonomous operation across lighting, weather, and seasonal change by adding colour-constant images \cite{Paton2015} and multi-experience localization \cite{Paton2016}. Multi-experience localization collects data every time the robot repeats a path and the most relevant experiences are chosen for feature matching. 

Developing a robust and accurate VT\&R system has taken a large research and engineering effort. As a result we can use outdoor datasets collected with VT\&R across lighting and seasonal change to compile multi-experience localization into a deep neural network (DNN) for relative pose estimation. VT\&R, which is shown to achieve high-accuracy path following \cite{Clement2016}, stores data in a spatio-temporal pose graph (see Figure \ref{fig:pose_graph}). The pose graph contains the relative pose between temporally adjacent keyframes derived from VO and the relative pose of a keyframe with respect to the mapped path. Each traversal of the path is stored as a new experience. We sample relative poses between keyframes that are localized across different experiences and use them as labels for our training data. We design the DNN based on previous work by Melekhov et al. \cite{Melekhov2017}. In particular, our DNN takes two pairs of stereo images and regresses the relative robot pose. In multi-experience localization VT\&R relies on gradually adding new experiences over time to be able to localize when the environment changes. We aim to localize radically different path traversals against each other without the use of such intermediate bridging experiences.   

\begin{figure}
    \centering
    \includegraphics[width=0.43\textwidth]{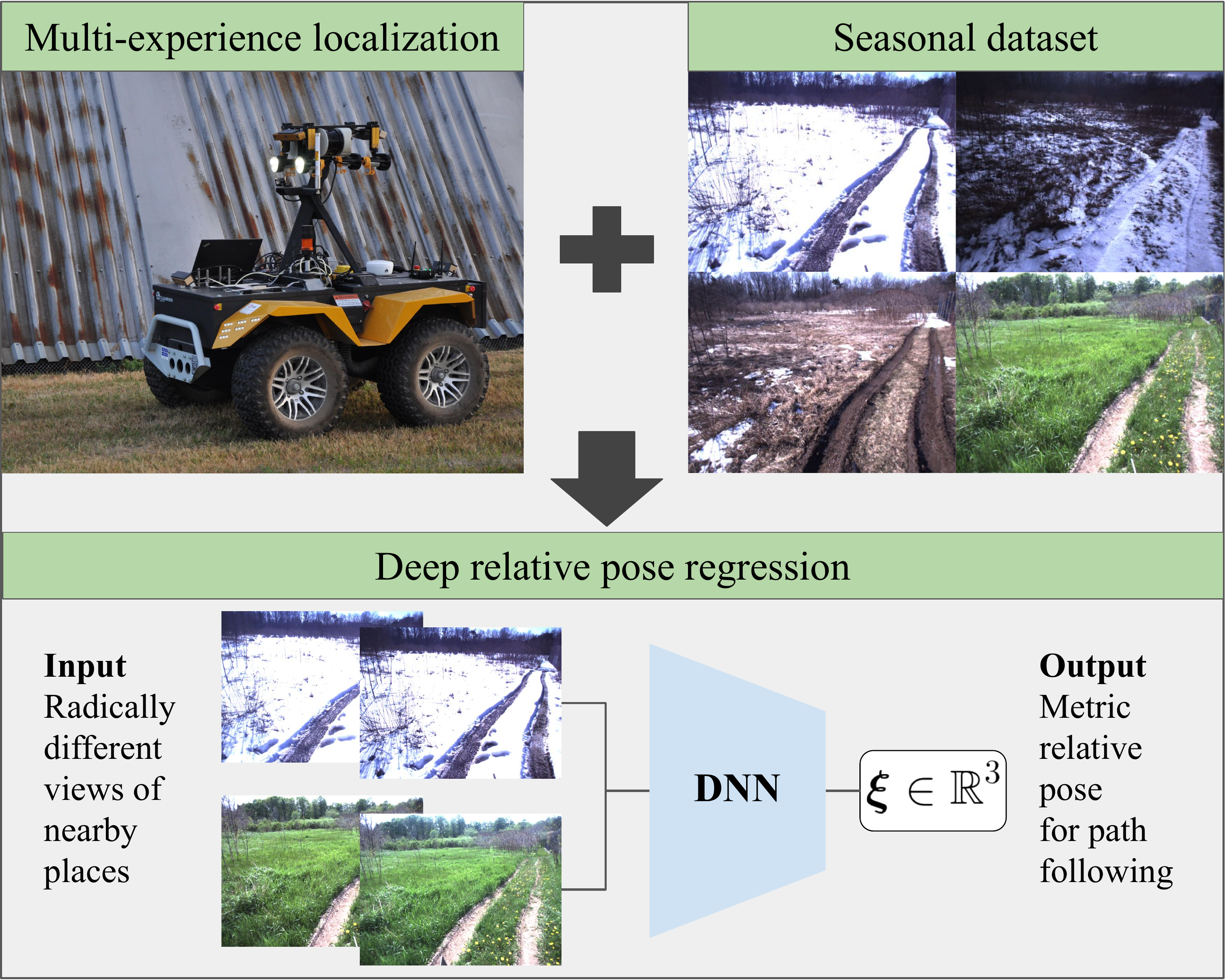}
    \caption{\footnotesize We compile multi-experience localization for path following into a DNN. We use datasets collected with VT\&R across different lighting and seasons to train the DNN to perform 3 degrees of freedom (DOF) relative pose estimation under changing environmental conditions.}
    \vspace{-0.6cm}
    \label{fig:overview}
\end{figure}


We conduct experiments to test the ability of our regressor to generalize across large appearance change. In VT\&R VO is used to propagate the current pose forward, while localization provides a pose correction by estimating the relative pose of the live frame with respect to the map. Since both VO and localization compute relative poses, we test our network's performance on both of these tasks. Using the exact same network architecture, we train one network with temporally adjacent keyframes for VO and one network with keyframes localized across different experiences.

The remainder of this paper is outlined as follows: Section \ref{sec:related_work} discusses related work, Section \ref{sec:methodology} gives the details of the network architecture and loss function, Section \ref{sec:experiments} explains our training procedure and lays out the experiments, while Section \ref{sec:results} provides the results.

\section{Related Work}
\label{sec:related_work}
VT\&R \cite{Furgale2010} performs accurate \cite{Clement2016} and robust autonomous path following. Moreover, the addition of colour-constant imagery \cite{Paton2015} and multi-experience localization \cite{Paton2016} enables the system to handle lighting, weather, and seasonal change.

Convolutional Neural Networks (CNN) have been included in different parts of the visual pose estimation pipeline to tackle appearance change. Several such approaches are tested against the Long-Term Visual Localization Benchmark \cite{sattler2018}. Examples include learning robust descriptors  \cite{Dymczyk2016, Sunderhauf2015b, sarlin2019, piasco2019, dusmanu2019, luo2019, revaud2019}, semantic information \cite{schonberger2018, larsson2019}, and place recognition \cite{arandjelovic2016}, and transforming whole images to different conditions \cite{Clement2018}. 

Others have in turn focused on replacing the whole pose-estimation pipeline with neural networks by regressing pose directly from images in an end-to-end fashion, several examples of which are presented in a survey on deep learning and visual simultaneous localization and mapping (SLAM) \cite{Li2018}. Early work on absolute pose regression came from the development of PoseNet \cite{Kendall2015}. The system is based on a pre-trained GoogleNet architecture and regresses 6-DOF pose for metric relocalization of a monocular camera. Kendall et al. extended the work to use a Bayesian neural network providing relocalization uncertainty \cite{Kendall2016} and an improved loss function \cite{Kendall2017}.  Naseer et al. \cite{Naseer2017} improve on PoseNet by generating additional augmented data leading to improved accuracy, while Walch et al. \cite{Walch2017} perform structured dimensionality reduction on the CNN output with the help of long short-term memory (LSTM) units. In \cite{Clark2017} and \cite{Patel2018} the authors were able to reduce localization error by passing sequential image data to recurrent models with LSTM units.    

Melekhov et al. \cite{Melekhov2017} use a Siamese CNN architecture based on AlexNet \cite{Krizhevsky2012} to compute relative camera pose from a pair of images. Similarly Bateux et al. \cite{Bateux2018} regress relative pose for use in visual servoing. VO is a special case of relative pose estimation, which has been explored extensively in the context of deep learning \cite{Mohanty2016, Wang2018, Zhan2018,Iyer2018, Costante2015}. In several examples authors combine CNNs with LSTM units to incorporate a sequence of data  \cite{Wang2018, Iyer2018}. Iyer et al. \cite{Iyer2018} use geometric consistency constraints to train their network in a self-supervised manner. In a different approach, Peretroukhin et. al. have combined deep learning with traditional pose estimation by learning pose corrections \cite{Peretroukhin2018} and  rotation \cite{Peretroukhin2019}, which they fuse with relative pose estimates.

Relative pose estimation has also been used as a tool to regress absolute pose. In particular, Laskar et al. \cite{Laskar2018} combine relative pose regression with image retrieval from a database. Balntas et al. \cite{Balntas2018} retrieve nearest neighbours based on learned image features before regressing relative pose to refine the absolute pose. Saha et al. \cite{Saha2018} classify anchor points to which they regress relative pose. Oliveira et al. \cite{Oliveira2017} combine the outputs of two DNNs for visual odometry and absolute pose estimation, respectively, to accomplish topometric localization. The work was further extended by using multi-task learning for localization \cite{Valada2018}. In our work we focus on robustness to large appearance change for relative pose regression and show experimental results on two challenging outdoor paths. 

\section{Methodology}
\label{sec:methodology}
\begin{figure}
    \centering
    \includegraphics[width=0.47\textwidth]{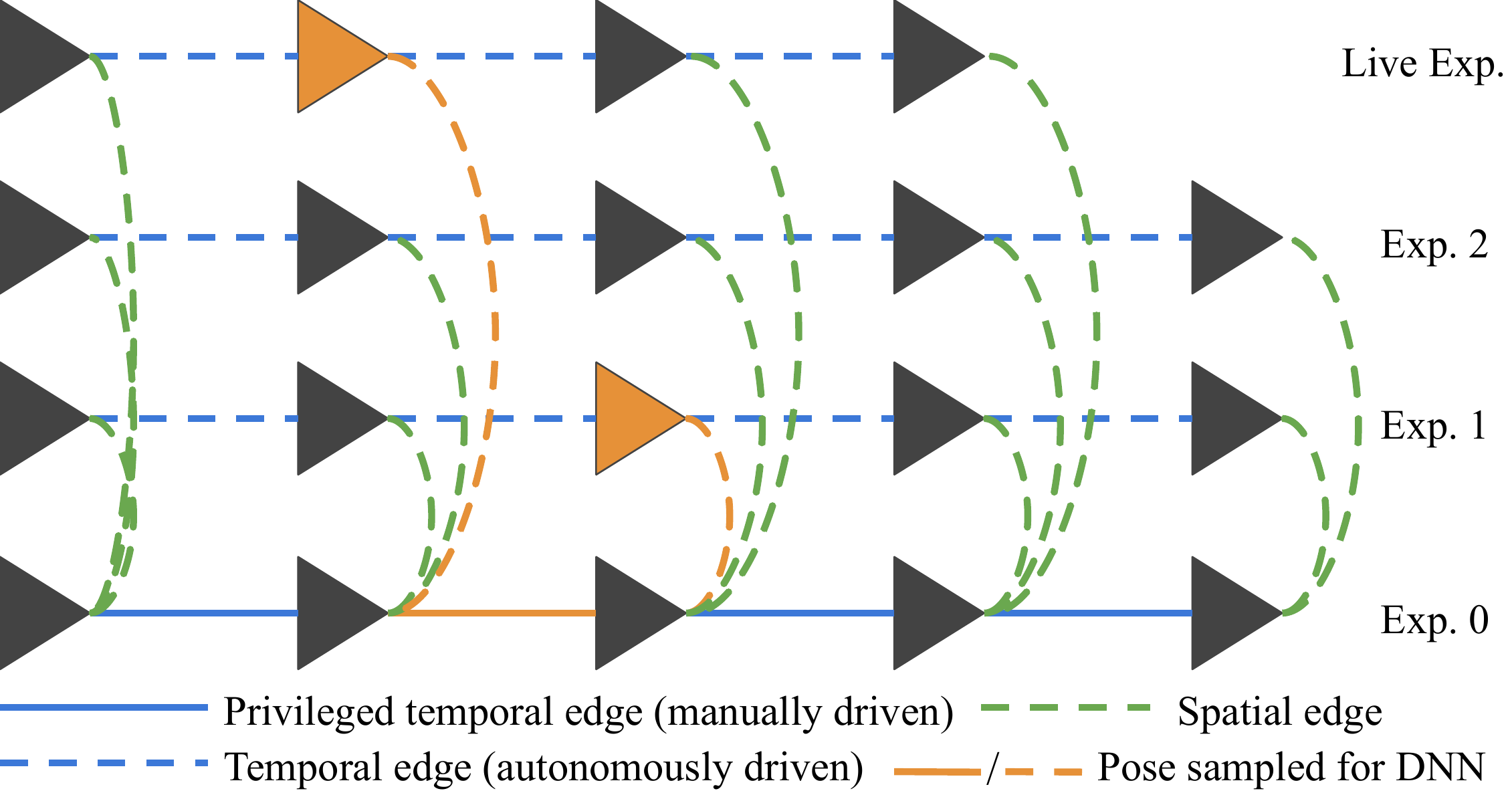}
    \caption{\footnotesize Image keyframes and the relative poses between them are stored in a spatio-temporal pose graph. Temporal edges represent relative poses from VO and spatial edges give the relative pose between a keyframe on a live experience and a keyframe on the privileged teach path. Several live keyframes may localize to the same privileged keyframe. The vertices and edges in orange show how we can sample relative poses in time and space, by compounding spatial and temporal transforms, to use as labels for training the DNN.}
    \label{fig:pose_graph}
    \vspace{-0.6cm}
\end{figure}
\begin{figure}
    \centering
    \includegraphics[width=0.44\textwidth]{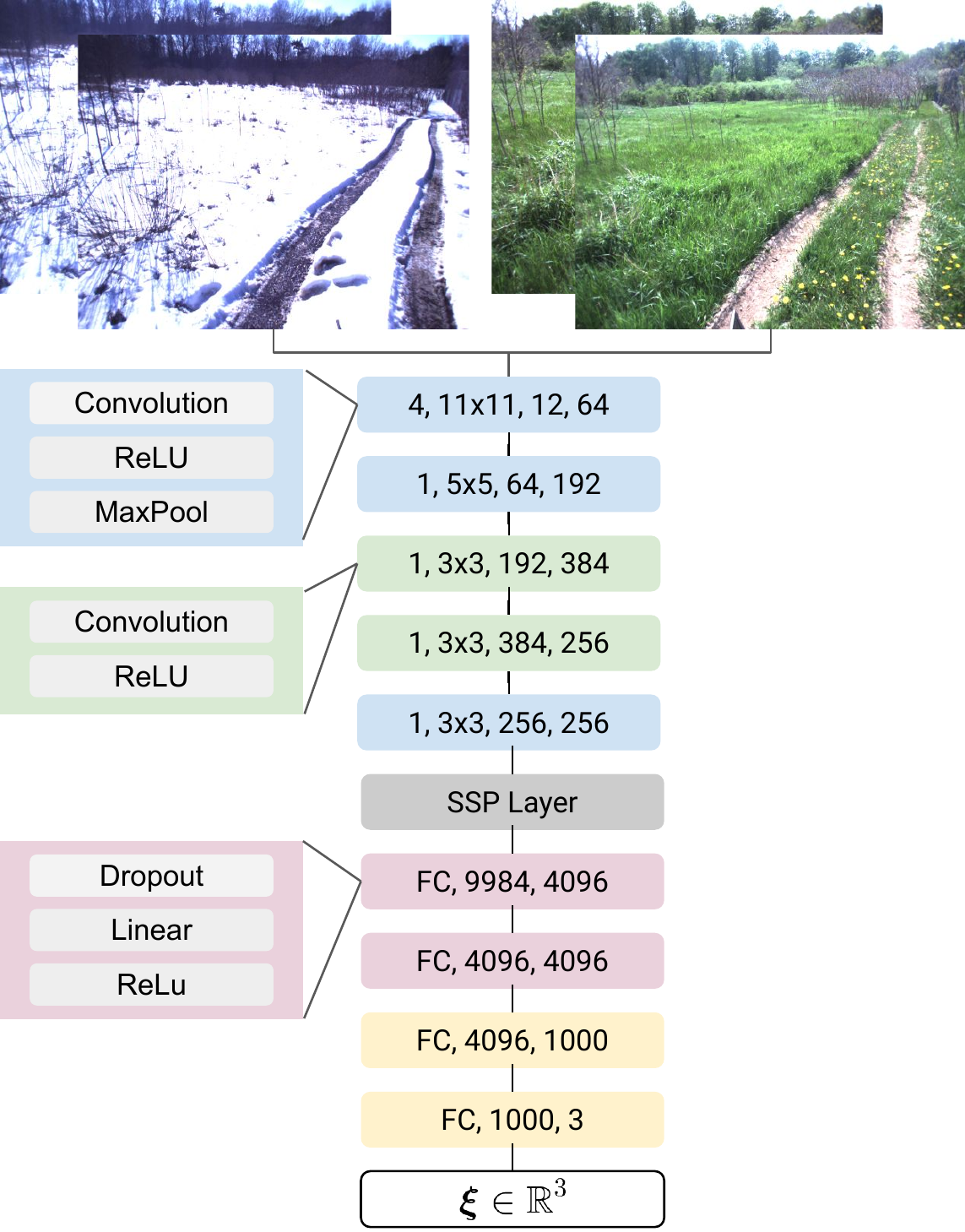}
    \caption{\footnotesize The neural network takes two sets of RGB stereo images and produces a 3-DOF relative pose. The architecture contains convolution layers, spatial pyramid pooling, and fully connected layers. We list the stride, kernel size, and number of input and output channels for the convolutional layers as well as input and output sizes for the fully connected layers.}
    \label{fig:network}
    \vspace{-0.3cm}
\end{figure}

\subsection{System Overview}
VT\&R stores image keyframes as vertices and relative poses between them as edges in a spatio-temporal pose graph. Figure \ref{fig:pose_graph} illustrates a pose graph with temporal edges derived from VO and spatial edges that connect keyframes from autonomous repeats to keyframes on the manually driven teach pass. Our system estimates both relative pose for VO as well as metric localization with respect to the path. We use the same neural network architecture and train two networks separately on data for VO and localization. Since VT\&R provides highly accurate path following \cite{Clement2016}, we sample relative pose labels from the VT\&R pose graph. The DNN takes as input RGB stereo images from a pair of keyframes and regresses a 3-DOF relative pose given in the robot frame. For path following the offset from the path and heading are the most important DOF and so we opt to estimate $\mbs{\xi} = \bbm x &  y & \theta \ebm ^T \in \Real^3$.

\subsection{Network Architecture}
Our DNN architecture is inspired by the one presented in \cite{Melekhov2017}. As in \cite{Melekhov2017}, the convolutional part of the DNN is taken from the AlexNet architecture \cite{Krizhevsky2012}. We opt to input a stack of all four RGB images to the network as in \cite{Wang2018}, resulting in 12 input channels. Experimenting with a Siamese architecture did not cause improvements in our case. Our images are different in size ($512 \times 384$) from the standard input to AlexNet ($224 \times 224$) and hence we make use of Spatial Pyramid Pooling (SPP) \cite{He2015} as in \cite{Melekhov2017} to reduce the size of our feature map before the fully connected layers. SPP lets us create a fixed-sized output while maintaining spatial information by pooling the responses of each feature in spatial bins (we use max pooling). The size of the output is the number of bins times the number of features. We use four levels of pyramid pooling with the following bins: $5 \times 5, 3 \times 3, 2 \times 2$, and $1 \times 1$. Finally, we keep the same fully connected layers as in AlexNet, but add one more fully connected layer with 3 connections to regress the 3-DOF pose. An overview of the network can be seen in Figure \ref{fig:network}. 

\subsection{Loss Function}

We use a simple quadratic loss function that takes the difference in target and predicted coordinates. Translation and rotation are manually weighted using a diagonal matrix $\mbf{W}$ with $1.0$ on the diagonal for $x$ and $y$ and $10.0$ for $\theta$. As pointed out in \cite{Kendall2017}, angles may wrap around $2\pi$, but this is not a problem we would encounter as we are estimating small relative poses. The loss is 
\begin{equation}
\mathcal{L} = \frac{1}{2}\left(\mbs{\xi} - \hat{\mbs{\xi}}\right)^T\mbf{W}\left(\mbs{\xi} - \hat{\mbs{\xi}}\right),
\end{equation}
where
$\mbs{\xi}$ represents the target pose that we have sampled from the VT\&R pose graph and $\hat{\mbs{\xi}}$ is the estimated pose.

\section{Experiments}
\label{sec:experiments}
\begin{figure}[t]
    \centering
    \begin{minipage}{0.24\textwidth}
        \centering
        \includegraphics[width=\textwidth]{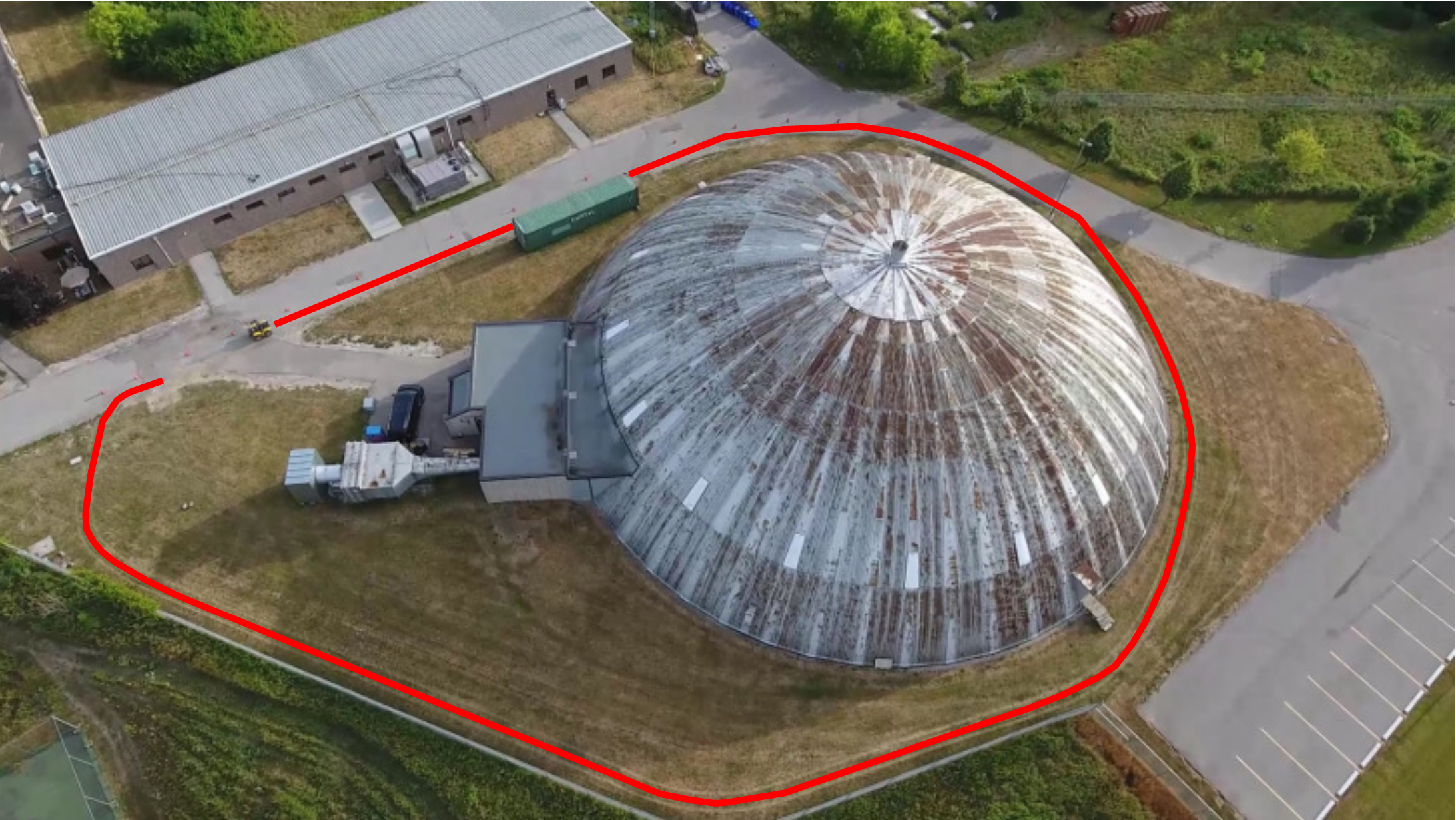}
    \end{minipage}%
    \hspace{1.00mm}
    \begin{minipage}{0.225\textwidth}
        \centering
        \includegraphics[width=\textwidth]{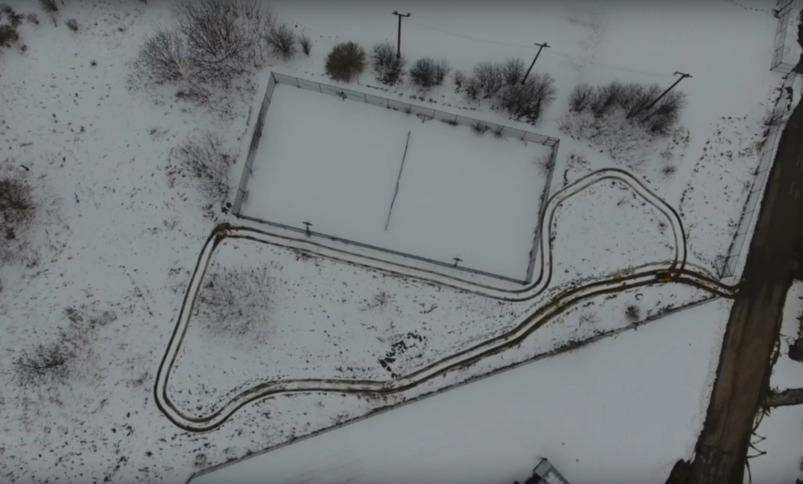}
    \end{minipage}%
    \caption{\footnotesize Aerial view of the paths for the UTIAS In The Dark and UTIAS Multi Season datasets.}
    \label{fig:paths}
    \vspace{-0.6cm}
\end{figure}
\begin{figure}[b]
    \centering
    \vspace{-0.5cm}
    \begin{minipage}{0.235\textwidth}
        \centering
        \includegraphics[width=\textwidth]{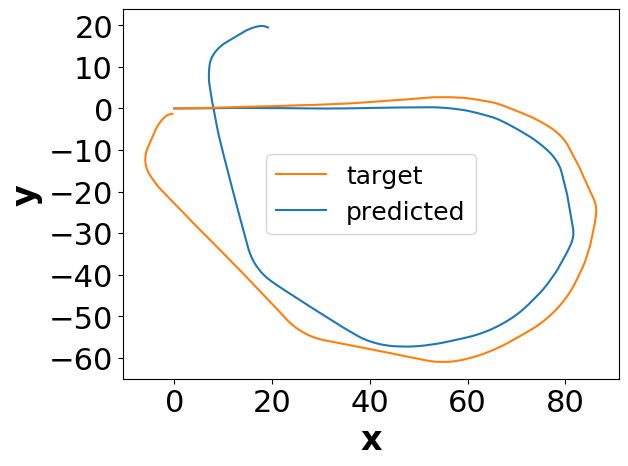}
    \end{minipage}%
    \hspace{1.00mm}
    \begin{minipage}{0.235\textwidth}
        \centering
        \includegraphics[width=\textwidth]{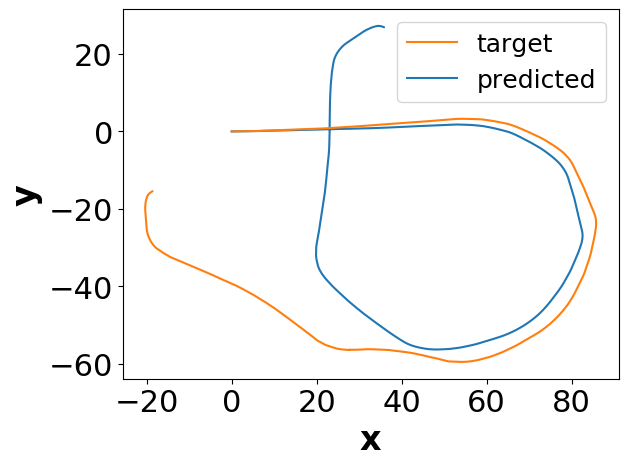}
    \end{minipage}%
    \caption{\footnotesize Integrated VO for day and evening representing one of the most and least accurate results, respectively.}
    \label{fig:vo_dark}
\end{figure}
\begin{figure}
 \vspace{-0.5cm}
    \centering
    \begin{minipage}{0.235\textwidth}
        \centering
        \includegraphics[width=\textwidth]{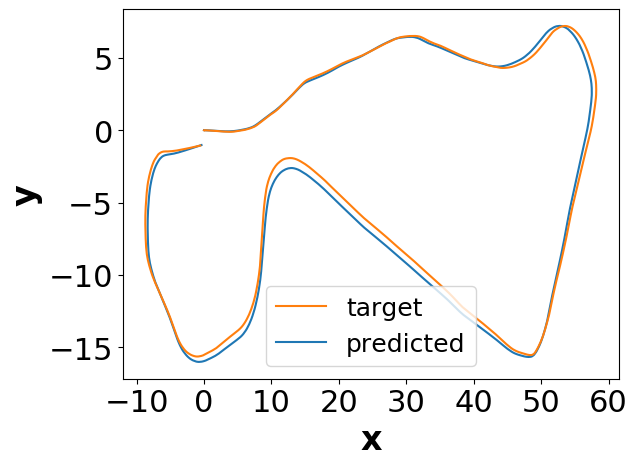}
    \end{minipage}%
    \hspace{1.00mm}
    \begin{minipage}{0.235\textwidth}
        \centering
        \includegraphics[width=\textwidth]{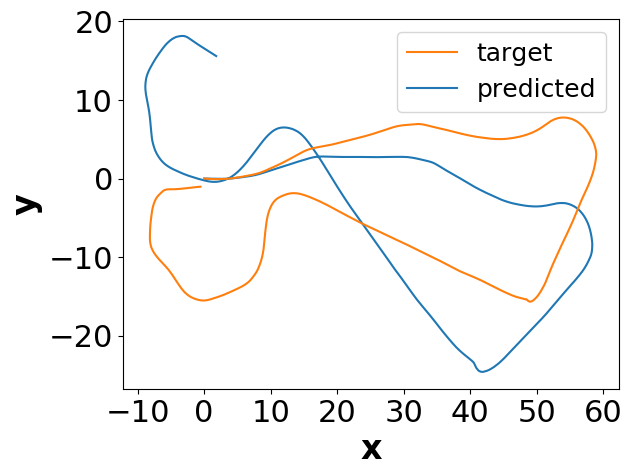}
    \end{minipage}%
    \caption{\footnotesize Integrated VO for sunny weather with snow on the ground and overcast weather with no snow representing one of the most and least accurate results, respectively.}
    \label{fig:vo_multiseason}
\end{figure}
\begin{table*}[t]
\centering
\caption{\footnotesize RMSE for each DOF for the UTIAS Multi Season dataset. The diagonal entries are VO results, while the off-diagonal entries are localization results. The rows are used as repeats and the columns as teach runs. The green and red cells are better and worse performing examples, respectively, picked for further qualitative analysis.}
\resizebox{0.8\textwidth}{!}{
\begin{tabular}{|c|c|l|l|l|l|l|l|l|l|}
\hline
\multicolumn{2}{|c|}{\multirow{2}{*}{}} 
& \multicolumn{2}{|c|}{Snow}                                                                                                                                                                                                                                                                                                                                                                      & \multicolumn{2}{|c|}{Some snow}                                                                                                                                                                                                                                                                                                                                                                 & \multicolumn{2}{|c|}{No snow}                                                                                                                                                                                                                                                                                                                                                                   & \multicolumn{2}{|c|}{Green}                                                                                                                                                                                                                                                                                                                                                                    \\ \hline
\multicolumn{2}{ |c| }{\multirow{2}{*}{}} & Sunny & Overcast & Sunny & Overcast & Sunny & Overcast & Sunny & Overcast \\ \hline

\multirow{2}{*}{Snow}      
& Sun                 	& \cellcolor{light_green} 
						  \begin{tabular}[l]{@{}l@{}} $x: 0.015 $\\ $y: 0.0039 $\\ $\theta: 0.11 $\end{tabular} 
  						& \begin{tabular}[l]{@{}l@{}} $x: 0.073 $\\ $y: 0.023 $\\ $\theta: 0.54 $\end{tabular} 
  						& \begin{tabular}[l]{@{}l@{}} $x: 0.086 $\\ $y: 0.028 $\\ $\theta: 0.55 $\end{tabular}
  						& \begin{tabular}[l]{@{}l@{}} $x: 0.086 $\\ $y: 0.041 $\\ $\theta: 0.85 $\end{tabular}
  						& \begin{tabular}[l]{@{}l@{}} $x: 0.070 $\\ $y: 0.017 $\\ $\theta: 0.40 $\end{tabular}
  						& \begin{tabular}[l]{@{}l@{}} $x: 0.075 $\\ $y: 0.023 $\\ $\theta: 0.49 $\end{tabular}
  						& \begin{tabular}[l]{@{}l@{}} $x: 0.089 $\\ $y: 0.028 $\\ $\theta: 0.59 $\end{tabular}
  						& \begin{tabular}[l]{@{}l@{}} $x: 0.082 $\\ $y: 0.024 $\\ $\theta: 0.49 $\end{tabular} \\ \cline{2-10}

& Overcast 				& \begin{tabular}[l]{@{}l@{}} $x: 0.073 $\\ $y: 0.031 $\\ $\theta: 0.57 $\end{tabular}
						& \begin{tabular}[l]{@{}l@{}} $x: 0.032 $\\ $y: 0.0032 $\\ $\theta: 0.11 $\end{tabular}
						& \begin{tabular}[l]{@{}l@{}} $x: 0.074 $\\ $y: 0.031 $\\ $\theta: 0.60 $\end{tabular}
						& \begin{tabular}[l]{@{}l@{}} $x: 0.088 $\\ $y: 0.046 $\\ $\theta: 0.81 $\end{tabular}
						& \begin{tabular}[l]{@{}l@{}} $x: 0.074 $\\ $y: 0.027 $\\ $\theta: 0.55 $\end{tabular}
						& \begin{tabular}[l]{@{}l@{}} $x: 0.080 $\\ $y: 0.037 $\\ $\theta: 0.59 $\end{tabular}
						& \begin{tabular}[l]{@{}l@{}} $x: 0.099 $\\ $y: 0.040 $\\ $\theta: 0.79 $\end{tabular}
						& \begin{tabular}[l]{@{}l@{}} $x: 0.088 $\\ $y: 0.035 $\\ $\theta: 0.68 $\end{tabular} \\ \hline
                           
\multirow{2}{*}{Some snow} 
& Sun      				& \begin{tabular}[l]{@{}l@{}} $x: 0.077 $\\ $y: 0.029 $\\ $\theta: 0.64 $\end{tabular}
						& \begin{tabular}[l]{@{}l@{}} $x: 0.075 $\\ $y: 0.032 $\\ $\theta: 0.65 $\end{tabular}
						& \begin{tabular}[l]{@{}l@{}} $x: 0.014 $\\ $y: 0.0059 $\\ $\theta: 0.13 $\end{tabular}
						& \begin{tabular}[l]{@{}l@{}} $x: 0.086 $\\ $y: 0.037 $\\ $\theta: 0.92 $\end{tabular}
						& \begin{tabular}[l]{@{}l@{}} $x: 0.070 $\\ $y: 0.028 $\\ $\theta: 0.61 $\end{tabular}
						& \begin{tabular}[l]{@{}l@{}} $x: 0.074 $\\ $y: 0.029 $\\ $\theta: 0.59 $\end{tabular}
						& \begin{tabular}[l]{@{}l@{}} $x: 0.100 $\\ $y: 0.039 $\\ $\theta: 0.81 $\end{tabular}
						& \begin{tabular}[l]{@{}l@{}} $x: 0.091 $\\ $y: 0.035 $\\ $\theta: 0.70 $\end{tabular} \\ \cline{2-10}

& Overcast 				& \begin{tabular}[l]{@{}l@{}} $x: 0.13 $\\ $y: 0.071 $\\ $\theta: 2.1 $\end{tabular}
						& \begin{tabular}[l]{@{}l@{}} $x: 0.13 $\\ $y: 0.069 $\\ $\theta: 2.1 $\end{tabular}
						& \begin{tabular}[l]{@{}l@{}} $x: 0.12 $\\ $y: 0.066 $\\ $\theta: 2.4 $\end{tabular}
						& \cellcolor{light_red}
						  \begin{tabular}[l]{@{}l@{}} $x: 0.019 $\\ $y: 0.0025 $\\ $\theta: 0.13 $\end{tabular}
						& \begin{tabular}[l]{@{}l@{}} $x: 0.12 $\\ $y: 0.065 $\\ $\theta: 1.3 $\end{tabular}
						& \begin{tabular}[l]{@{}l@{}} $x: 0.13 $\\ $y: 0.069 $\\ $\theta: 2.1 $\end{tabular}
						& \begin{tabular}[l]{@{}l@{}} $x: 0.12 $\\ $y: 0.071 $\\ $\theta: 1.4 $\end{tabular}
						& \cellcolor{light_red}
						  \begin{tabular}[l]{@{}l@{}} $x: 0.13 $\\ $y: 0.076 $\\ $\theta: 1.6 $\end{tabular}\\ \hline
                                                      
\multirow{2}{*}{No snow}   
& Sun			        & \cellcolor{light_green}
						  \begin{tabular}[l]{@{}l@{}} $x: 0.056 $\\ $y: 0.017 $\\ $\theta: 0.39 $\end{tabular}
						& \begin{tabular}[l]{@{}l@{}} $x: 0.061 $\\ $y: 0.020 $\\ $\theta: 0.41 $\end{tabular}
						& \begin{tabular}[l]{@{}l@{}} $x: 0.065 $\\ $y: 0.023 $\\ $\theta: 0.48 $\end{tabular}
						& \begin{tabular}[l]{@{}l@{}} $x: 0.079 $\\ $y: 0.031 $\\ $\theta: 0.63 $\end{tabular}
						& \begin{tabular}[l]{@{}l@{}} $x: 0.011 $\\ $y: 0.0033 $\\ $\theta: 0.10 $\end{tabular}
						& \begin{tabular}[l]{@{}l@{}} $x: 0.057 $\\ $y: 0.021 $\\ $\theta: 0.42 $\end{tabular}
						& \begin{tabular}[l]{@{}l@{}} $x: 0.076 $\\ $y: 0.025 $\\ $\theta: 0.52 $\end{tabular}
						& \begin{tabular}[l]{@{}l@{}} $x: 0.074 $\\ $y: 0.019 $\\ $\theta: 0.46 $\end{tabular} \\ \cline{2-10}

& Overcast 				& \begin{tabular}[l]{@{}l@{}} $x: 0.071 $\\ $y: 0.025 $\\ $\theta: 0.49 $\end{tabular}
						& \begin{tabular}[l]{@{}l@{}} $x: 0.067 $\\ $y: 0.034 $\\ $\theta: 0.47 $\end{tabular}
						& \begin{tabular}[l]{@{}l@{}} $x: 0.070 $\\ $y: 0.028 $\\ $\theta: 0.57 $\end{tabular}
						& \begin{tabular}[l]{@{}l@{}} $x: 0.082 $\\ $y: 0.033 $\\ $\theta: 0.75 $\end{tabular}
						& \begin{tabular}[l]{@{}l@{}} $x: 0.057 $\\ $y: 0.024 $\\ $\theta: 0.47 $\end{tabular}
						& \begin{tabular}[l]{@{}l@{}} $x: 0.012 $\\ $y: 0.0042 $\\ $\theta: 0.012 $\end{tabular}
						& \begin{tabular}[l]{@{}l@{}} $x: 0.082 $\\ $y: 0.031 $\\ $\theta: 0.61 $\end{tabular}
						& \begin{tabular}[l]{@{}l@{}} $x: 0.074 $\\ $y: 0.028 $\\ $\theta: 0.54 $\end{tabular} \\ \hline
                           
\multirow{2}{*}{Green}     
& Sun           		& \begin{tabular}[l]{@{}l@{}} $x: 0.097 $\\ $y: 0.034 $\\ $\theta: 0.63 $\end{tabular}
						& \begin{tabular}[l]{@{}l@{}} $x: 0.10 $\\ $y: 0.039 $\\ $\theta: 0.92 $\end{tabular}
						& \begin{tabular}[l]{@{}l@{}} $x: 0.10 $\\ $y: 0.042 $\\ $\theta: 0.81 $\end{tabular}
						& \begin{tabular}[l]{@{}l@{}} $x: 0.095 $\\ $y: 0.045 $\\ $\theta: 0.96 $\end{tabular}
						& \begin{tabular}[l]{@{}l@{}} $x: 0.088 $\\ $y: 0.031 $\\ $\theta: 0.66 $\end{tabular}
						& \begin{tabular}[l]{@{}l@{}} $x: 0.097 $\\ $y: 0.036 $\\ $\theta: 0.74 $\end{tabular}
						& \begin{tabular}[l]{@{}l@{}} $x: 0.019 $\\ $y: 0.0033 $\\ $\theta: 0.14 $\end{tabular}
						& \begin{tabular}[l]{@{}l@{}} $x: 0.070 $\\ $y: 0.029 $\\ $\theta: 0.50 $\end{tabular} \\ \cline{2-10}

& Overcast 				& \begin{tabular}[l]{@{}l@{}} $x: 0.090 $\\ $y: 0.029 $\\ $\theta: 0.55 $\end{tabular}
						& \begin{tabular}[l]{@{}l@{}} $x: 0.099 $\\ $y: 0.032 $\\ $\theta: 0.69 $\end{tabular}
						& \begin{tabular}[l]{@{}l@{}} $x: 0.10 $\\ $y: 0.033 $\\ $\theta: 0.65 $\end{tabular}
						& \begin{tabular}[l]{@{}l@{}} $x: 0.10 $\\ $y: 0.042 $\\ $\theta: 0.94 $\end{tabular}
						& \begin{tabular}[l]{@{}l@{}} $x: 0.10 $\\ $y: 0.026 $\\ $\theta: 0.52 $\end{tabular}
						& \begin{tabular}[l]{@{}l@{}} $x: 0.097 $\\ $y: 0.030 $\\ $\theta: 0.61 $\end{tabular}
						& \begin{tabular}[l]{@{}l@{}} $x: 0.089 $\\ $y: 0.030 $\\ $\theta: 0.52 $\end{tabular}
						& \begin{tabular}[l]{@{}l@{}} $x: 0.013 $\\ $y: 0.0035 $\\ $\theta: 0.14 $\end{tabular} \\ \hline
\end{tabular}}
\label{tab:multiseason}
\vspace{-0.6cm}
\end{table*}

\begin{table}
\centering
\caption{\footnotesize RMSE for each DOF for the UTIAS In The Dark dataset. The diagonal entries are VO results, while the off-diagonal entries are localization results. The rows are used as repeats and the columns as teach runs. The green and red cells are better and worse performing examples, respectively, picked for further qualitative analysis.}
\resizebox{\columnwidth}{!}{
\begin{tabular}{|c|c|c|c|c|c|}
\hline
& Morning & Sun Flare & Day & Evening & Night \\ \hline

Morning    	& \begin{tabular}[l]{@{}l@{}} $x: 0.0073 $\\ $y: 0.0022 $\\ $\theta: 0.080 $\end{tabular}
			& \begin{tabular}[l]{@{}l@{}} $x: 0.012 $\\ $y: 0.0053 $\\ $\theta: 0.17 $\end{tabular}
			& \begin{tabular}[l]{@{}l@{}} $x: 0.013 $\\ $y: 0.0058 $\\ $\theta: 0.15 $\end{tabular}
			& \begin{tabular}[l]{@{}l@{}} $x: 0.013 $\\ $y: 0.0079 $\\ $\theta: 0.15 $\end{tabular}
			& \begin{tabular}[l]{@{}l@{}} $x: 0.013 $\\ $y: 0.0093 $\\ $\theta: 0.21 $\end{tabular} \\ \hline

Sun Flare	& \cellcolor{light_green}
              \begin{tabular}[l]{@{}l@{}} $x: 0.010 $\\ $y: 0.0051 $\\ $\theta: 0.12 $\end{tabular}
			& \begin{tabular}[l]{@{}l@{}} $x: 0.0079 $\\ $y: 0.0023 $\\ $\theta: 0.084 $\end{tabular}
			& \begin{tabular}[l]{@{}l@{}} $x: 0.011 $\\ $y: 0.0060 $\\ $\theta: 0.14 $\end{tabular}
			& \begin{tabular}[l]{@{}l@{}} $x: 0.013 $\\ $y: 0.0069 $\\ $\theta: 0.15 $\end{tabular}
			& \begin{tabular}[l]{@{}l@{}} $x: 0.013 $\\ $y: 0.0086 $\\ $\theta: 0.20 $\end{tabular}\\ \hline
                           
Day        	& \begin{tabular}[l]{@{}l@{}} $x: 0.012 $\\ $y: 0.0058 $\\ $\theta: 0.13 $\end{tabular}
           	& \begin{tabular}[l]{@{}l@{}} $x: 0.011 $\\ $y: 0.0058 $\\ $\theta: 0.14 $\end{tabular}
     		& \cellcolor{light_green} 
     		  \begin{tabular}[l]{@{}l@{}} $x: 0.0074 $\\ $y: 0.0021 $\\ $\theta: 0.079 $\end{tabular}
     		& \begin{tabular}[l]{@{}l@{}} $x: 0.013 $\\ $y: 0.0069 $\\ $\theta: 0.15 $\end{tabular}
     		& \begin{tabular}[l]{@{}l@{}} $x: 0.013 $\\ $y: 0.0087 $\\ $\theta: 0.20 $\end{tabular} \\ \hline

Evening     & \begin{tabular}[l]{@{}l@{}} $x: 0.019 $\\ $y: 0.011 $\\ $\theta: 0.22 $\end{tabular}
			& \begin{tabular}[l]{@{}l@{}} $x: 0.019 $\\ $y: 0.011 $\\ $\theta: 0.22 $\end{tabular}
			& \begin{tabular}[l]{@{}l@{}} $x: 0.019 $\\ $y: 0.011 $\\ $\theta: 0.22 $\end{tabular}
			& \cellcolor{light_red}
			  \begin{tabular}[l]{@{}l@{}} $x: 0.0091 $\\ $y: 0.0037 $\\ $\theta: 0.092 $\end{tabular}
			& \begin{tabular}[l]{@{}l@{}} $x: 0.020 $\\ $y: 0.012 $\\ $\theta: 0.28 $\end{tabular} \\ \hline

Night       & \begin{tabular}[l]{@{}l@{}} $x: 0.015 $\\ $y: 0.013 $\\ $\theta: 0.29 $\end{tabular}
            & \begin{tabular}[l]{@{}l@{}} $x: 0.015 $\\ $y: 0.014 $\\ $\theta: 0.31 $\end{tabular}
            & \begin{tabular}[l]{@{}l@{}} $x: 0.016 $\\ $y: 0.013 $\\ $\theta: 0.29 $\end{tabular}
            & \cellcolor{light_red}
              \begin{tabular}[l]{@{}l@{}} $x: 0.016 $\\ $y: 0.014 $\\ $\theta: 0.31 $\end{tabular}
            & \begin{tabular}[l]{@{}l@{}} $x: 0.0047 $\\ $y: 0.0041 $\\ $\theta: 0.092 $\end{tabular} \\ \hline
                           
\end{tabular}}
\vspace{-0.7cm}
\label{tab:dark}
\end{table}
We conduct experiments to test relative pose estimation for VO and localization, where localization is performed between stereo camera frames taken during different times of day, weather, and seasons. The experiments make use of data collected with a Clearpath Grizzly RUV with a maximum speed of 1 m/s equipped with a factory-calibrated PointGrey Bumblebee XB3 stereo camera with 24 cm baseline, see Figure \ref{fig:overview}. We use VT\&R with colour-constant images \cite{Paton2015} and multi-experience localization \cite{Paton2016} to label the data. Multi-experience localization stores each traversal of the path in the pose graph. During a repeat a set of the experiences most similar to the current conditions are chosen for feature matching when localizing with respect to the path. 

\subsection{Training and Testing}

We train, validate, and test our system on two outdoor paths. The first dataset, called UTIAS In The Dark, covers a 250 m path following a paved road and grass in an area with buildings. The path is repeated once per hour for over 24 hours covering significant lighting change. The robot has headlights for driving during the night. The path has 45 repeats from which we choose 5 for testing and use the remaining for training and validation. We only train and test our network once for each path and do not re-train for each test condition. The path in the second dataset, called UTIAS Multi Season, is about 160 m. It covers an area with rugged terrain and vegetation. Data is collected from winter to spring and includes a total of 138 repeats, 8 of which are held out for testing. Figure \ref{fig:paths} shows an aerial view of the paths.  

Data collected during path traversals by the hand-engineered VT\&R system is organized in a spatio-temporal pose graph illustrated in Figure \ref{fig:pose_graph}. With the help of multi-experience localization, VT\&R is able to localize back to the teach pass during long-term driving, providing us with training data across large environmental change. In order to generate pose labels for training and validation for VO, we sample the temporal edges between immediately adjacent keyframes. For localization we can sample randomly from the graph in both space and time, allowing us to generate large datasets connecting keyframes from both similar and radically different conditions. An example of such a sample is illustrated in orange in Figure \ref{fig:pose_graph}. We randomly pick a vertex from an autonomous repeat and find to which privileged teach keyframe this vertex is localized. If we want to sample in space we can pick another teach keyframe in the same area. Finally, we randomly pick an autonomous repeat vertex localized to the chosen teach vertex. We compound the transforms to get the relative pose associated with our pair of keyframes. For this paper we sample only in time and do not move along the graph in the spatial direction. For the UTIAS In The Dark dataset our training and validation sets have $360,000$ and $40,000$ samples for localization, respectively. For VO we get $64,530$ and $7170$ samples. The UTIAS Multi Season dataset has $450,000$ and $50,000$ samples for localization training and validation, respectively. For VO we have $69,659$ training samples and $7739$ validation samples.

When processing the test runs we keep the data sequential to assess performance in a realistic scenario. We test localization across environmental change by performing localization for every pair of runs in the test set (one run is used as the teach pass and the other as the autonomous repeat). Specifically, this lets us localize radically different traversals directly without the use of any intermediate bridging experiences. We test VO standalone for the same runs. 

Path following is performed by alternating between using VO to propagate the pose forward and localization to provide a pose correction. We perform two experiments for localization. In order to test the localization network standalone we compute the relative pose between the live and teach keyframes that are localized to each other by VT\&R and compare directly to the VT\&R labels. Note that VT\&R does not consider global pose estimates, only the relative pose of the robot with respect to the teach pass. We also include a qualitative path following experiment, where we test the VO and localization networks together. We start by computing the relative pose between the initial live and teach keyframe pair. Next we use VO, as computed by the network, to propagate this pose forward for a window of possible next teach keyframes and choose the teach keyframe that gives the smallest new relative pose. As the correction step we use the localization network to compute a relative pose between the live and teach keyframes. We combine the propagated pose with the pose from localization by computing a weighted average with weights 0.3 and 0.7, respectively.     

We train our network on an NVIDIA GTX 1080 Ti GPU with a batch size of 64 and use early stopping based on the validation loss to determine the number of epochs. We use the Adam optimizer \cite{Kingma2014} with learning rate 0.0001 and other parameters set to their default values. Network inference runs at a minimum 50 fps on a Lenovo laptop with one GPU.

\section{Results}
\label{sec:results}

\begin{figure*}
    \centering
    \begin{minipage}{0.94\textwidth}
        \includegraphics[width=\textwidth]{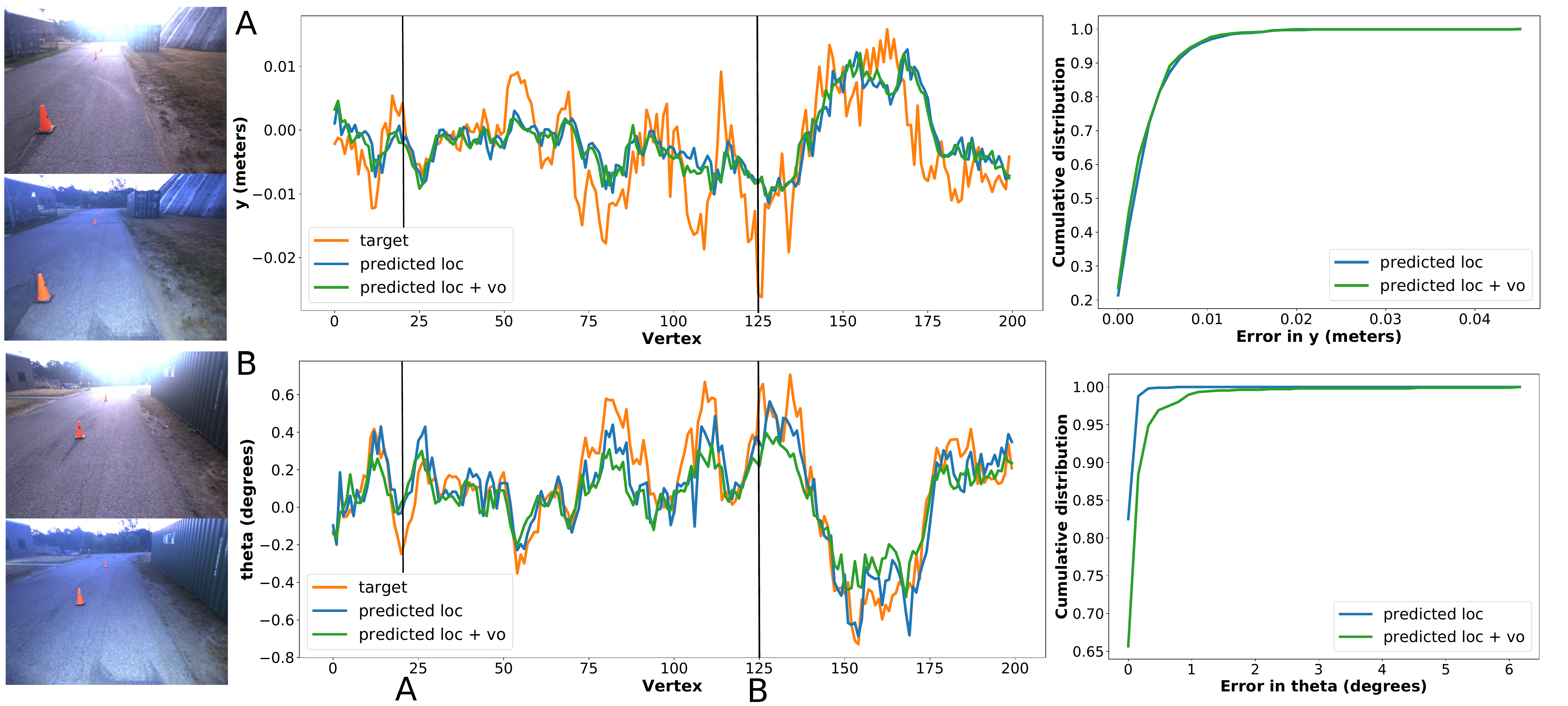}
    \end{minipage}
    \begin{minipage}{0.94\textwidth}
        \includegraphics[width=\textwidth]{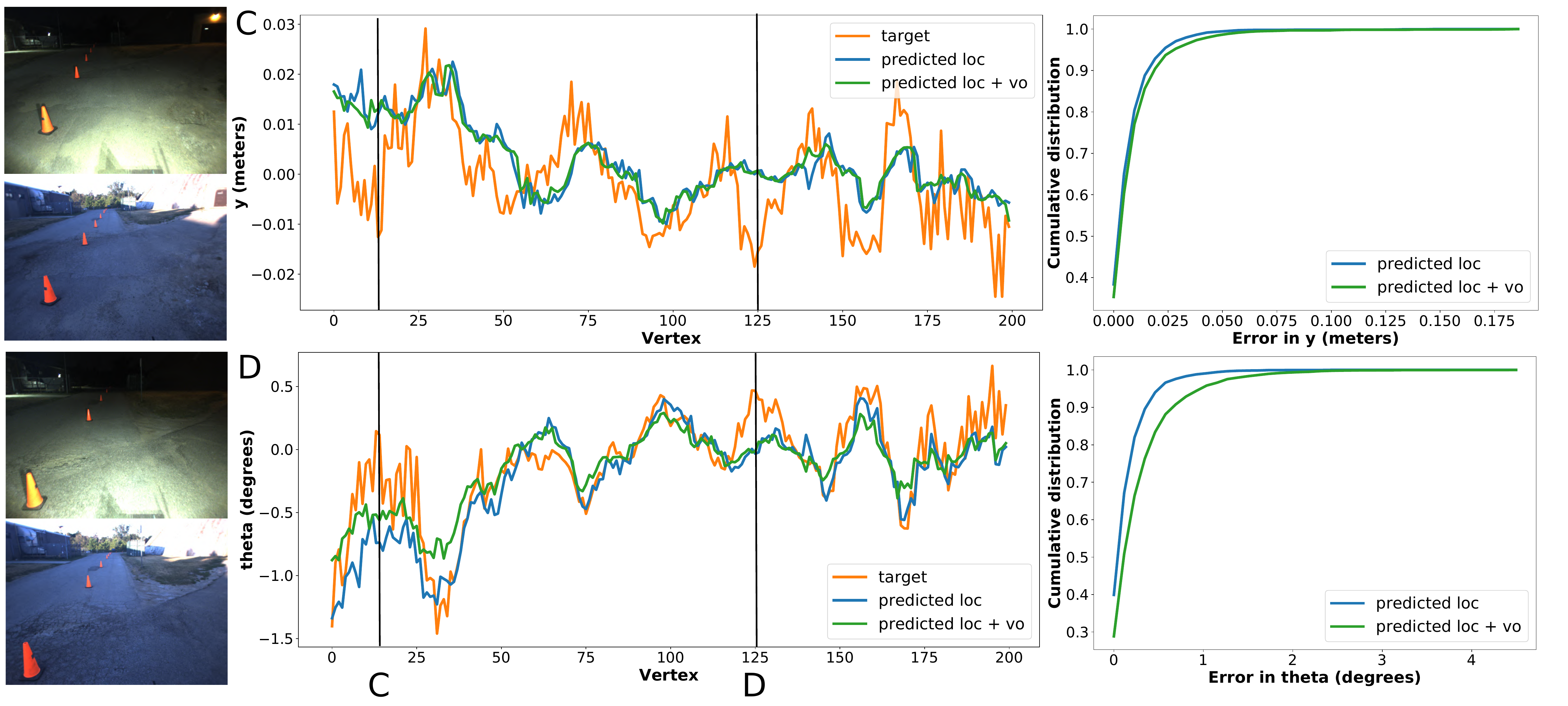}
    \end{minipage}
    \caption{\footnotesize The figure shows localization results for one of the better (top) test sequences from the UTIAS In The Dark dataset and one with larger errors (bottom). We plot the relative pose estimates for $y$ and $\theta$ from the localization network (blue) as well as pose estimates from combining VO and localization (green) together with the target values for a segment of the full path. The plots on the right show the cumulative distribution of errors for the full path. The image pairs on the left provide anecdotal examples from the test sequence and are marked in the plot.}
    \label{fig:loc_dark}
    \vspace{-0.7cm}
\end{figure*}
\begin{figure*}
    \centering
    \begin{minipage}{0.94\textwidth}
        \includegraphics[width=\textwidth]{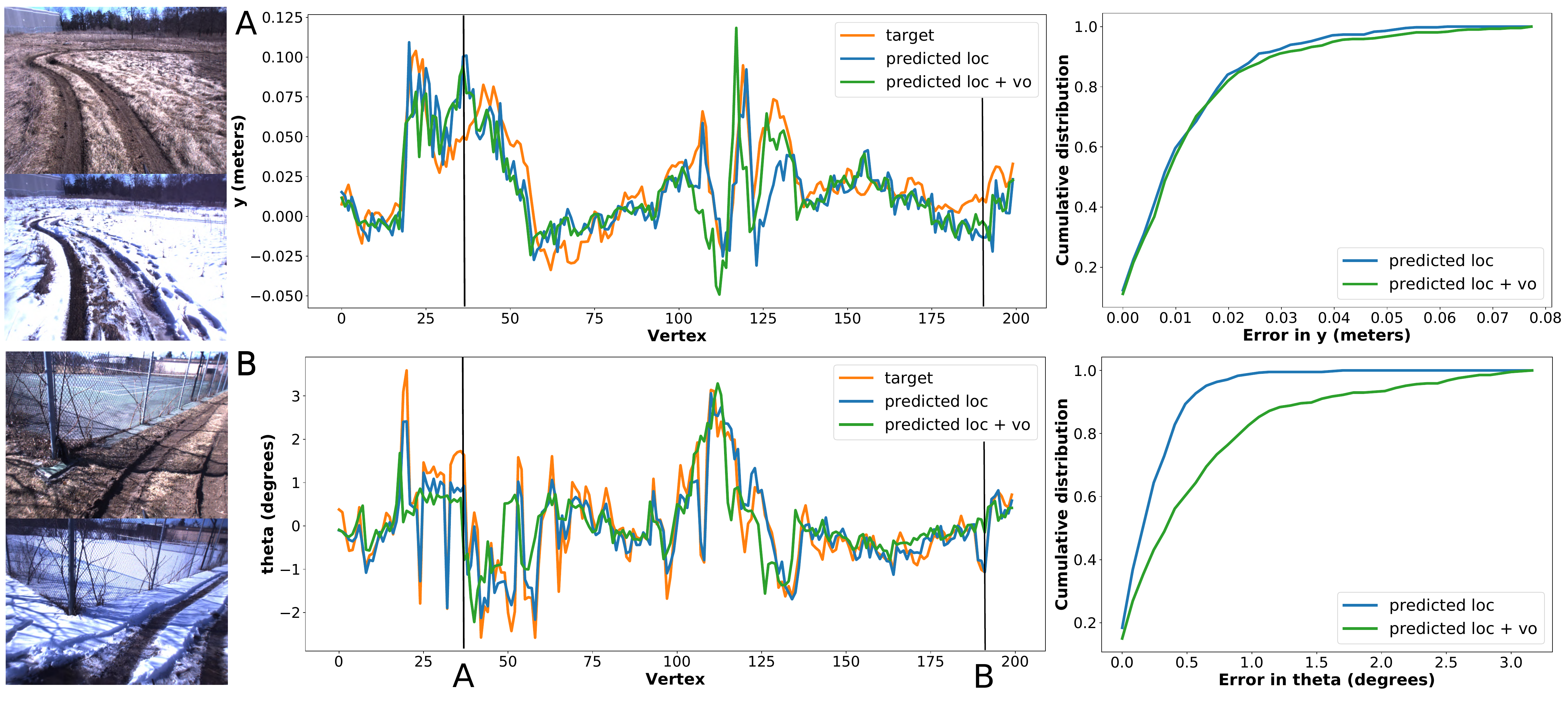}
    \end{minipage}
    \begin{minipage}{0.94\textwidth}
        \includegraphics[width=\textwidth]{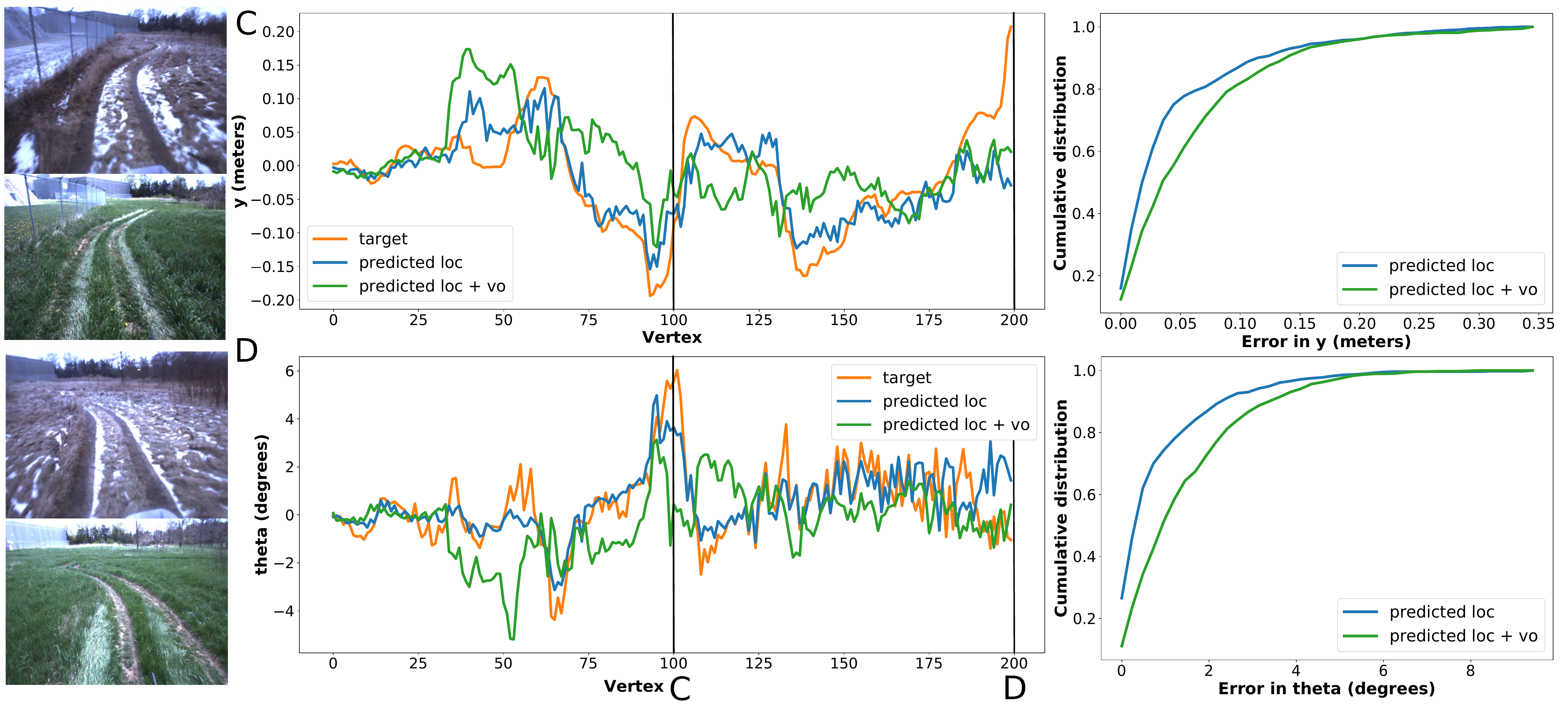}
    \end{minipage}
    \caption{\footnotesize The figure shows localization results for one of the better (top) test sequences from the UTIAS Multi Season dataset and one with larger errors (bottom). We plot the relative pose estimates for $y$ and $\theta$ from the localization network (blue) as well as pose estimates from combining VO and localization (green) together with the target values for a segment of the full path. The plots on the right show the cumulative distribution of errors for the full path. The image pairs on the left provide anecdotal examples from the test sequence and are marked in the plot.}
    \label{fig:loc_multiseason}
    \vspace{-0.7cm}
\end{figure*}

We conduct standalone experiments for the localization and VO networks for data with large appearance variation in an outdoor environment. Tables \ref{tab:multiseason} and \ref{tab:dark} list the root mean squared error (RSME) for each run in the test sets. We compare performance with the VT\&R system, which we know has centimeter-level error on kilometer-scale repeats \cite{Clement2016}. The values on the diagonal are results for VO, while the rest are for localization. The rows represent repeat runs while the columns are used as teach runs. For the paths in our datasets the the relative pose values for $x$ can typically fall between 0 to 30 cm. $y$ normally varies between +/- 10 cm, but can reach almost 40 cm on sharp turns. Similarly $\theta$ mostly varies between +/- 5 degrees, but may reach almost 40 degrees on sharp turns. If the network had only learned to randomly return small pose estimates, path following would not be possible due to the difference in relative pose size on straight road versus turns. Furthermore, repeat speed and the number of repeat keyframes can vary between runs and so simply replaying previous experiences would also fail quickly. With these approximate numbers in mind, we see that our system achieves low errors across a range of conditions. For tests across lighting change localizing evening and night repeats are the most challenging, but they do not perform much worse than the other combinations. For the seasonal tests we see that the network is able to localize runs as different as winter and spring. These examples show the system's potential to localize against large environmental changes directly without relying on intermediate experiences.

To supplement our quantitative findings we provide more detailed plots for two test cases from each dataset. We pick one of the best performing test cases (marked in green in the tables) and one of the cases with the largest errors (marked in red). We integrate the results from VO to show the full paths in Figures \ref{fig:vo_dark} and \ref{fig:vo_multiseason}, while Figures \ref{fig:loc_dark} and \ref{fig:loc_multiseason} display localization results. For path following, the most important performance indicators are the lateral and heading errors with respect to the path. For a small segment of each path we plot the target $y$ and $\theta$ values against those predicted standalone by the localization network as well as the path following method that combines VO and localization network outputs for prediction and correction. We think the latter method has a smoothing effect on the pose estimates. The fact that this method chooses different teach keyframes for localization than the original VT\&R system may account for some of the discrepancy between the two solutions in Figure \ref{fig:loc_multiseason}. Additionally, we plot the cumulative distribution of $y$ and $\theta$ errors for the whole test sequence and include two example teach-and-repeat image pairs, illustrating the challenging environmental change. The path from the UTIAS In The Dark Dataset has less sharp turns and smaller lateral path offsets resulting in a smaller signal-to-noise ratio in the data making the value of $y$ harder to predict, see Figure \ref{fig:loc_dark}. Given the RMS errors as well as the plots from the example runs, we think that this localization system would be sufficiently accurate for path following in the loop. 

\section{Conclusion and Future Work}
\label{sec:conclusion}
In this work, we present a DNN that can perform relative pose regression for both VO and localization with respect to a path in an outdoor environment across illumination and seasonal change. We collect labels for training and testing from a spatio-temporal pose graph generated by VT\&R. We conduct experiments across environmental change on two outdoor paths. The network carries out VO under different and challenging conditions, including night time driving. Furthermore, our network can perform localization for input image pairs from different times of day or  seasons without the need of intermediate bridging experiences, which are necessary for long-term operation with the original VT\&R system. From the performance we achieve on these datasets we believe that the localization system is sufficiently accurate for in-the-loop path following.

Tackling the localization problem across outdoor environmental change is a first step in applying deep learning more generally to path following. We want to improve the technique by enabling transfer to paths not seen during training. Ultimately we aim to close the loop in real time with the localization system we have developed in this paper.



\section*{ACKNOWLEDGMENT}

We would like to thank Clearpath Robotics and the Natural Sciences and Engineering Research Council of Canada (NSERC) for supporting this work.



\clearpage
\newpage
\bibliographystyle{IEEEtran}
\bibliography{ms.bbl}

\end{document}